# Beyond Churn: Predicting Financial Fragmentation in Retail Banking with Temporal Machine Learning

Ananyaa Chopra, Brandon Xu, Brendan Yuen, Lauren Zung, Sarabroop Aulakh
reachananyaa@gmail.com, xubrandon89@gmail.com, brendanyuen@hotmail.com, laurenzung@gmail.com, sarabroopsinghaulakh@gmail.com

Royal Bank of Canada

**Abstract**

Retail banking attrition is usually represented as a terminal binary event, even though client relationships often weaken earlier through partial movements of deposits, investments, and recurring activity to external financial institutions. This paper defines that preceding state as *financial fragmentation* and presents an end-to-end temporal machine-learning system for predicting it before complete disengagement. Using anonymized multi-source data from a large retail bank, the framework predicts whether a valid external transfer or investment event will occur within 90 days. The study uses 595,220 client-month observations, with 346 engineered features combining monthly client profiles, balances, product relationships, prior flow-of-funds behavior, macroeconomic conditions, and competitor activity. A four-stage XGBoost cascade estimates (1) whether an external outflow will occur within 90 days, (2) the expected amount, (3) the originating product, and (4) the destination financial institution. The primary classifier achieved a test precision-recall area under the curve of 0.823. At the validation-selected threshold, it produced 86.4% precision, 75.1% recall, and an F1 score of 0.803. Ranking test observations in descending Stage 1 fragmentation score, the top 1% of clients yielded 95.3% precision, while the top 5% captured 78.7% of observed outflow cases. The amount model placed 94.9% of predictions within an adjacent amount bucket. Destination prediction reached a macro-F1 of 0.81 across 27 classes; source-product prediction achieved a weighted F1 of 0.92. By moving the analytical focus from terminal churn to earlier fund migration, the proposed approach provides a practical foundation for proactive, explainable, and economically informed client-retention decision support.

## 1 Introduction

Retail banking relationships rarely disappear at a single observable moment. A client may keep a chequing account open while moving savings to a digital bank, transferring investments to a self-directed platform, or redirecting recurring financial activity to a specialist provider. From a conventional churn perspective, that client remains active. From an economic and relationship perspective, however, the institution may already be losing share of wallet, future fee income, and the opportunity to serve the client's evolving needs.

This gap matters because most churn models are designed around an end state: closure, prolonged inactivity, or formal defection. Prior research has shown that the definition of churn materially changes both the target population and the apparent predictive performance of a model (Neslin et al., 2006). Research on partial defection similarly demonstrates that customers can reduce the depth of a multi-product relationship without leaving completely (Ahn et al., 2006). In retail banking, where clients can hold products across several institutions simultaneously, the boundary between retained and lost is therefore not binary.

We define **financial fragmentation** as the redistribution of a client's financial assets or transactional activity across external institutions while the primary banking relationship remains open. Here, 'primary' relationship refers to the institution whose active client relationship is being modeled; it does not require that the institution still hold the majority of the client's products, assets, or liquidity. The definition is intentionally behavioural rather than motivational: an external transfer does not necessarily indicate dissatisfaction, nor does every movement warrant intervention. It does, however, create an observable event that can be predicted, quantified, and investigated. This reframing moves the analytical question from *Who has already left?* to *Whose relationship is beginning to fragment, how much value may move, from which product, and toward which institution?*

The project addresses that question through a four-stage predictive decision-intelligence system. Stage 1 estimates the probability of at least one qualifying external flow-of-funds event within the next 90 days. Sequential models then estimate the likely outflow amount, originating product, and destination financial institution. These outputs are combined into an expected-outflow score that ranks opportunities by both likelihood and financial magnitude. The result is a decision-support layer designed to help portfolio, analytics, and relationship teams investigate risk, understand the behaviors associated with a prediction, and allocate limited outreach capacity more deliberately.

The setting creates several machine-learning challenges. Positive fragmentation events represent a small portion of client-month observations, creating a class imbalanced learning problem where the model must reliably detect rare positive cases against a dominant negative majority. This imbalance increases the importance of precision-recall analysis (Davis & Goadrich, 2006; Saito & Rehmsmeier, 2015). Repeated monthly observations introduce temporal dependence, while the 90-day forward label creates potential for inflated model confidence if features are highly correlated with future performance, though this risk is mitigated by careful feature selection using only sufficiently historical data. Monetary outcomes are heavily right-skewed, and originating products are dominated by chequing transactions. Finally, the use of predictions in a client-facing context requires transparent reasoning, restricted access, human review, and careful distinction between model association and causal explanation.

Gradient-boosted trees are well suited to this environment because they capture nonlinear interactions in heterogeneous tabular data, handle sparse and differently scaled inputs, and provide a practical balance between predictive strength and operational feasibility (Chen & Guestrin, 2016; Friedman, 2001). Large-scale benchmarking continues to find tree ensembles challenging to outperform on many irregular tabular problems (Grinsztajn et al., 2022). The present work therefore centers on XGBoost, which offer distinct advantages over neural architectures for structured financial data—specifically, interpretability of feature importance for regulatory compliance, robustness to missing values and imbalanced classes without extensive preprocessing, computational efficiency in both training and inference, and strong performance with limited labeled data.

This paper makes four contributions. First, it operationalizes financial fragmentation as a measurable pre-churn outcome with a defined observation date and 90-day horizon. Second, it presents a robust client-month pipeline that joins internal relationships and transaction data with external economic and competitive context. Third, it evaluates a cascading XGBoost system using rare-event, capacity-constrained, monetary, and multiclass measures. Fourth, it translates the model into an explainable, human-reviewed workflow through a continuous expected-outflow ranking and an interactive dashboard and conversational analysis layer.

The empirical results are promising but intentionally presented with boundaries. The Stage 1 classifier delivers strong precision and lift; the destination model performs well across a comparatively large class set, and the amount model captures broad order-of-magnitude differences. Conversely, some outflows remain difficult to estimate; minority source products are poorly recovered relative to chequing, and these data limitations shape the interpretation of the results and the next stage of development.

# 2 Related Work

## 2.1 Churn, partial defection, and banking relationships

Customer churn prediction has long been used to identify relationships at risk of termination. Previous work emphasizes that predictive conclusions depend heavily on how churn is defined, which observations are included, and how performance is measured (Neslin et al., 2006). In multi-product markets, complete defection is only one possible outcome: customers may reduce usage, move selected products, or maintain a low-value residual relationship. Ahn, Han, and Lee's analysis of partial defection is especially relevant because it treats relationship erosion as a process rather than a single endpoint (Ahn et al., 2006).

Recent banking research has applied a wide range of supervised methods to customer attrition. In a study conducted on a large bank, multiple machine-learning approaches using rich behavioral data showed the value of prior client activity for predicting future churn (Lima Lemos et al., 2022). Other work has compared ensemble and neural approaches across standard churn settings (Lalwani et al., 2022). These studies establish the practical relevance of behavioral machine learning in financial institutions, but most published formulations retain a static client record and a binary terminal target.

Financial fragmentation differs in two ways. First, the event is an observed external movement rather than account closure. Second, the target is in a defined future window relative to a monthly snapshot. A client labeled as non-fragmenting in one month may fragment in the next, allowing the model to learn within-client behavioral shifts in addition to between-client heterogeneity. This formulation is closer to early-warning event prediction than conventional one-record-per-customer churn.

## 2.2 Value-aware and cost-sensitive retention

Predictive accuracy alone does not determine whether a retention system creates value. A false positive consumes outreach capacity, a false negative may represent a missed retention opportunity, and the economic importance of two true positives can differ by orders of magnitude. Profit-driven churn research has therefore argued for objectives and rankings that incorporate treatment cost, customer value, and expected benefit (Verbeke et al., 2012). Cost-sensitive learning provides the broader theoretical foundation for aligning errors with their unequal consequences (Elkan, 2001).

The present framework follows that logic by separating *probability of fragmentation* from *magnitude of potential outflow*. A client with a 60% likelihood of moving a large balance may deserve earlier investigation than a client with a 95% likelihood of moving a small amount. The implemented score is deliberately simple and auditable: predicted probability multiplied by predicted amount, which generates a ranking of scores suited for prioritization. It can later be extended with an estimated treatment response and intervention cost, but those quantities are not assumed in the current results.

## 2.3 Gradient boosting for heterogeneous tabular data

Gradient boosting constructs an additive model by fitting successive weak learners to reduce a differentiable loss (Friedman, 2001). XGBoost adds a regularized objective, sparse-aware tree construction, and scalable implementation (Chen & Guestrin, 2016). Alternative systems such as LightGBM and CatBoost offer different efficiency and categorical-feature strategies (Ke et al., 2017; Prokhorenkova et al., 2018), while neural approaches such as TabNet and transformer-based tabular models have expanded the design space (Arik & Pfister, 2021; Gorishniy et al., 2021). Nevertheless, broad empirical comparisons show that tree-based ensembles remain strong baselines and often the leading methods on medium-sized tabular data with irregular feature distributions (Grinsztajn et al., 2022).

These properties align with the project data: balances, counts, ratios, lags, categorical encodings, and macroeconomic variables interact nonlinearly; missingness may itself carry signal; and model training must remain feasible within an enterprise development environment. XGBoost is well-suited to these constraints: it efficiently learns nonlinear interactions, handles missing data natively, and trains quickly on CPUs with modest computational overhead. Additionally, XGBoost provides early stopping, sample weighting for class imbalance, multiclass objectives, and built-in feature importance estimation, all within a single unified framework—reducing implementation complexity and development time.

## 2.4 Rare events and evaluation under limited capacity

Imbalanced learning research distinguishes between techniques that alter the training distribution and metrics that reveal minority-class performance (Chawla et al., 2002; He & Garcia, 2009). Receiver operating characteristic curves can appear favorable when there is a class imbalance, because the false-positive rate is normalized by a very large negative class, masking a high count of erroneous predictions that lower precision. Precision-recall analysis is more directly sensitive to the prevalence and retrieval quality of the positive class (Davis & Goadrich, 2006; Saito & Rehmsmeier, 2015).

For an operational retention workflow, rank-based measures are equally important. Portfolio teams rarely act on every score; they review the highest-risk fraction permitted by capacity. Precision at the top 1%, capture at the top 5%, and lift over the base rate therefore answer a more concrete question: *How concentrated are actual outflows among the clients that can realistically be reviewed?* The present evaluation emphasizes PR-AUC, precision, recall, F1, capture, and lift for this reason.

### 2.5 Temporal validation, model risk, and interpretability

Temporal prediction systems can overstate performance when observations from the future influence training or when random splits place closely related records on both sides of an evaluation boundary. Research on time-respecting evaluation and concept drift emphasizes that deployed performance depends on both strict temporal separation and robustness to changing market conditions (Bergmeir et al., 2018; Cerqueira et al., 2020; Gama et al., 2014). The project employs a three-month temporal buffer between the training period and held-out evaluation pool to prevent leakage. Validation and test were further separated within the July-September period to ensure no information from the test set influenced hyperparameter tuning.

Interpretability is also necessary but must be framed carefully. SHAP provides an additive framework for decomposing individual predictions and has efficient implementations for tree ensembles (Lundberg et al., 2020; Lundberg & Lee, 2017). The current project report uses XGBoost feature-importance scores to summarize leading predictors. These scores describe how the model uses available features; they do not establish that changing a feature will causally prevent fragmentation.

Finally, production machine learning creates risks beyond predictive error. Hidden technical debt can arise from unstable data dependencies, unmonitored drift, tightly coupled pipelines, and unclear ownership (Sculley et al., 2015). Model cards and dataset documentation provide useful principles for recording intended use, limitations, evaluation context, and governance (Gebru et al., 2021; Mitchell et al., 2019). Those principles inform the system design described below.

## 3 Methodology

### 3.1 Research setting and data sources

The analysis was conducted on a protected enterprise virtual machine using internal retail-banking data. The data contains eligible client snapshots per month after cohort and data-quality filters.

Four source families were joined at a monthly analytical grain.

**Table 1. Data sources and governance**

| Source | Grain | Principal content | Frequency and governance |
|---|---|---|---|
| Unified Client Profile | Client-month | Demographics, balances, product holdings, segmentation, relationship attributes | Monthly snapshot; personally identifiable data encrypted at rest and protected by role-based access |
| Flow-of-Funds transactions | Transaction event | Amount, date, originating product, destination account and institution | Event ingestion aggregated to client-month; sensitive financial data |
| Macroeconomic data | Calendar month | GDP, consumer prices, Bank of Canada rate, equity-market returns, unemployment | Monthly public-source data; no client identifiers |
| Competitor-offer data | Calendar month | Promotional counts and competitor rate updates | Internally maintained monthly/quarterly series |

The analytical unit is a **client-month snapshot**. Let $i$ index clients and $t$ denote the end of an observation month. The dataset is

$$\mathcal{D} = \{(\mathbf{x}_{i,t}, y_{i,t}, a_{i,t}, s_{i,t}, z_{i,t})\},$$

where $\mathbf{x}_{i,t} \in \mathbb{R}^{346}$ contains information available by the snapshot date, $y_{i,t}$ is the 90-day fragmentation label, $a_{i,t}$ is future outflow amount, $s_{i,t}$ is the originating-product class, and $z_{i,t}$ is the destination-institution class.

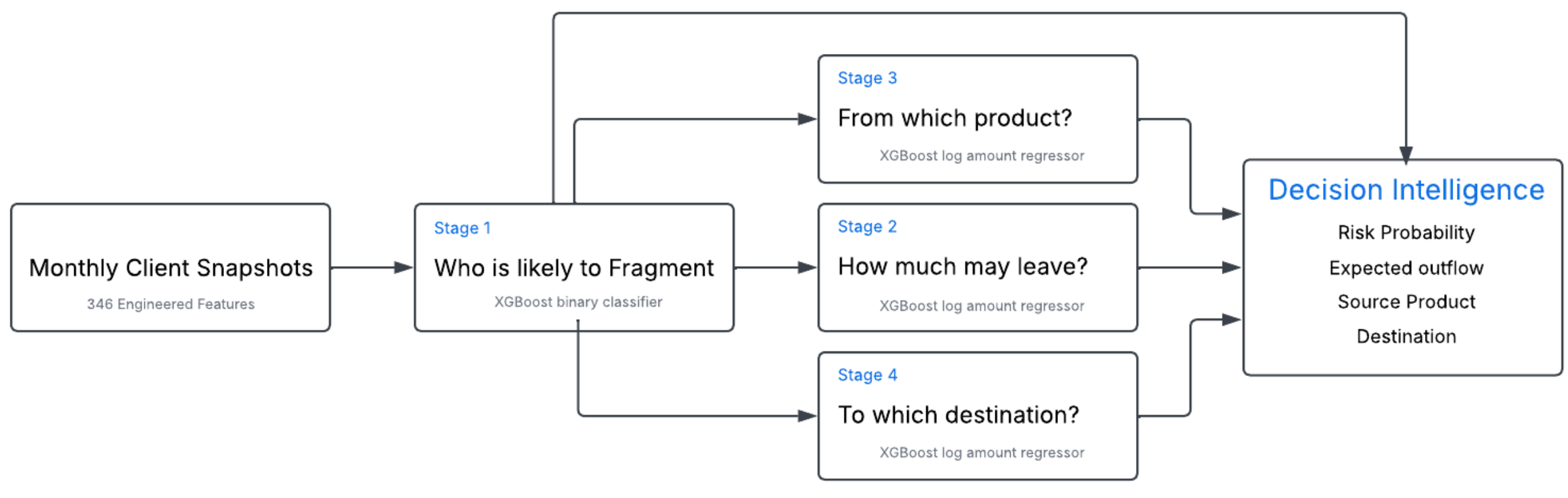

Priority Score: $\widehat{p_i} \times \widehat{A_i}$

The System rank outreach opportunities, it doesnt automate client action

The four-stage model converts a monthly client snapshot into a risk probability, amount estimate, source-product prediction, destination prediction, and human-reviewed priority list.

## 3.2 Cohort construction and temporal design

The final dataset contains 595,220 client-month observations. Training uses January-March 2025. April-June 2025 is excluded as a buffer to reduce overlap between historical training signals and the held-out evaluation window. July-September 2025 forms the evaluation pool and is divided approximately 50/50 at random into validation and test subsets. The validation set supports hyperparameter and threshold selection; the test set is not used for those choices.

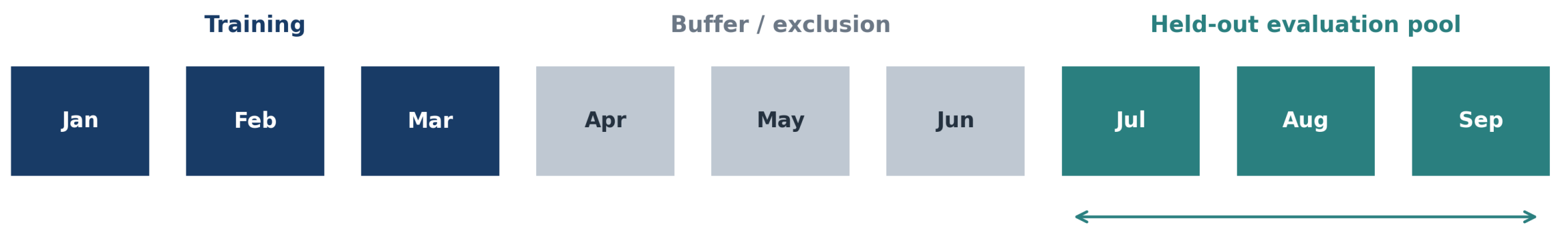

The design prevents leakage from the training period, but validation and test are not chronologically ordered relative to each other.

Training is separated from the evaluation pool by a three-month buffer. Validation and test are held out from training but are randomly divided from the same July-September period. This design is stronger than a random split across the full year.

## 3.3 Fragmentation label and conditional targets

A qualifying fragmentation event is an external transaction whose normalized destination product is either **OFI Investments** or **OFI Transfers** and whose destination financial institution is valid, external, and non-null. Internal bank destinations, unknown values, and not-applicable labels are excluded from label and destination-propensity construction. The primary label imposes no minimum dollar threshold; it is designed to detect the onset of external activity, including small early movements.

For a 90-day horizon $H$, define the future event set

$$\mathcal{F}_{i,t}^{(H)} = \{e : i_e = i,\ t < \tau_e \leq t + H,\ e \text{ is a valid external OFI event}\}.$$

The binary outcome is

$$y_{i,t} = \mathbb{1}\left(\left|\mathcal{F}_{i,t}^{(90)}\right| > 0\right),$$

and the cumulative amount target is

$$a_{i,t} = \sum_{e \in \mathcal{F}_{i,t}^{(90)}} v_e,$$

where $v_e$ is the absolute value of event $e$. The originating product $s_{i,t}$ is consolidated from 16 raw labels into six business-relevant groups: chequing, savings, mutual funds, wealth management, guaranteed investment certificates (GICs), and other. Destination strings are normalized to institution umbrellas; sparse institutions are grouped into other_fi, leaving 27 classes.

Stages 2-4 are trained exclusively on observations with actual positive outflow events, since amount, source, and destination features are defined only when fragmentation occurs. In operational deployment, Stage 1 serves as a practical gate: while the batch pipeline may compute downstream estimates for all rows to simplify data infrastructure, the dashboard surfaces Stages 2-4 outputs only for clients flagged by Stage 1. This gating reduces computational overhead and focuses analyst review on high-risk clients, rather than surfacing conditional predictions for the entire portfolio.

## 3.4 Feature engineering

The final feature matrix contains 346 predictors. All features are constructed using information available at or before the snapshot date. The principal families are shown in Table 2.

**Table 2. Feature families and leakage controls**

| Family | Representative variables | Temporal control |
|---|---|---|
| Client and relationship | Age band, income, tenure, segment, digital status, product count | Month-end snapshot only |
| Portfolio composition | Total, liquid, deposit, mutual-fund, GIC, registered-savings balances; logs and ratios | As of snapshot date or lagged |
| Behavioral activity | Inflow/outflow counts, amounts, product use, prior external activity | Historical windows end at $t$ |
| Lag/Slope dynamics | One-, two-, and three-month lags; changes; acceleration; decline flags | No transaction after $t$ enters a feature |
| Destination propensity | Distinct destination institutions, destination concentration, historical shares | Historical receiving institutions only |
| External context | Interest rates, inflation, unemployment, market returns, promotional counts | Joined by publication month |

Missing numerical values are mapped to zero only when zero has a valid business interpretation, such as no recorded balance. Categorical missingness is represented explicitly as unknown. Highly granular product and institution labels are consolidated before modeling to reduce extreme sparsity. The current development pipeline does not rely on personally identifying fields as model inputs.

## 3.5 Stage 1: 90-day fragmentation classifier

Stage 1 uses an XGBoost binary classifier with logistic loss. Let the ensemble prediction be

$$\hat{p}_i = \sigma\left(\sum_{k=1}^{K} f_k(\mathbf{x}_i)\right),$$

where $f_k$ is a regression tree and $\sigma(u) = (1 + e^{-u})^{-1}$. XGBoost minimizes a regularized objective of the form

$$\mathcal{L} = \sum_{i=1}^{N} w_i \, \ell(y_i, \hat{p}_i) + \sum_{k=1}^{K} \left(\gamma T_k + \frac{\lambda}{2} \parallel \boldsymbol{\omega}_k \parallel_2^2\right),$$

where $T_k$ is the number of leaves, $\boldsymbol{\omega}_k$ contains leaf weights, and $w_i$ is a client-value sample weight. The project uses total balance, $B_i$, as the sample weight to reflect economic significance—giving greater influence to higher-AUM clients without allowing a small number of outlier ultra-high-net-worth clients to dominate training. This ensures the model learns patterns that matter for portfolio risk while remaining robust to extreme cases:

$$\widetilde{w}_i = \max\{1, \log(1 + \max(B_i, 0))\}, \qquad w_i = \frac{\widetilde{w}_i}{N^{-1}\sum_{j=1}^{N}\widetilde{w}_j}.$$

The normalized training weights have mean 1.000 and range from 0.133 to 2.146 after normalization. A 40-trial randomized search evaluated class weighting (scale_pos_weight), tree depth, and regularization using validation PR-AUC and early stopping. No negative-class downsampling was applied; the selected scale_pos_weight of 1 emerged from hyperparameter tuning, suggesting that sample weighting by client balance was sufficient to address class imbalance without additional resampling.

**Table 3. Selected Stage 1 configuration**

| Parameter | Selected value | Rationale |
|---|---|---|
| scale_pos_weight | 1 | Preserve natural class balance; rely on evaluation and value weights rather than artificial positive upweighting |
| max_depth | 4 | Limit tree complexity and overfitting |
| learning_rate | 0.10 | Converge within the early-stopping budget |
| min_child_weight | 20 | Require support before creating a leaf |
| reg_lambda | 1.0 | L2 regularization |
| reg_alpha | 0.5 | L1 regularization |
| Best iteration | 158 of 2,000 | Validation early stopping |

The classification threshold is selected on validation data to maximize F1. This produces $\tau^* = 0.3321$ with validation precision 0.8657, recall 0.7567, and F1 0.8075.

### 3.6 Stage 2: conditional amount model

Future outflow amounts are highly right skewed. Stage 2 therefore models

$$r_i = \log(1 + a_i)$$

with an XGBoost regressor and transforms predictions back using $\hat{a}_i = \exp(\hat{r}_i) - 1$. The continuous target preserves ordering and provides a dollar estimate, while bucket-based weights increase attention to larger events without assigning unbounded influence:

$$q(a_i) = \begin{cases} 1.0, & 0 \le a_i < 500, \\ 1.5, & 500 \le a_i < 2{,}500, \\ 2.5, & 2{,}500 \le a_i < 12{,}500, \\ 4.0, & a_i \ge 12{,}500. \end{cases}$$

The final amount model uses max_depth=5, learning_rate=0.02, min_child_weight=50, subsample=0.7, colsample_bytree=0.7, reg_lambda=10.0, and reg_alpha=2.0. Model selection uses a composite of validation log-scale $R^2$ and recall for outflows above $12,500, balancing average fit against high-value detection.

### 3.7 Stages 3 and 4: originating product and destination

The originating-product and destination models use XGBoost's multiclass soft-probability objective. For class $c$,

$$P(Y_i = c \mid \mathbf{x}_i) = \frac{\exp(g_c(\mathbf{x}_i))}{\sum_{j=1}^{C}\exp\left(g_j(\mathbf{x}_i)\right)}.$$

Both models combine the total-balance sample weight with a square-root-dampened inverse-frequency class weight:

$$w_{i,c} = w_i\sqrt{\frac{N}{C\,n_c}},$$

where $n_c$ is the training count for class $c$. This reduces majority-class dominance while avoiding the extreme weights produced by a full inverse-frequency rule.

Both stages use n_estimators=500, max_depth=4, learning_rate=0.05, min_child_weight=20, subsample=0.7, colsample_bytree=0.7, reg_lambda=3.0, reg_alpha=1.0, early_stopping_rounds=25, and random_state=42.

### 3.8 Decision score and system integration

The operational ranking joins Stage 1 and Stage 2 through

$$\hat{E}_i = \hat{p}_i \times \hat{a}_i,$$

where $\hat{E}_i$ is expected outflow. This continuous score ranks clients by the product of likelihood and estimated financial magnitude. For example, a client with a 90% risk score and a predicted $50,000 outflow receives an expected-outflow score of $45,000, which ranks above a 95% risk client with a predicted $5,000 outflow.

A future retention-value formulation could incorporate estimated intervention response $r_i$ and cost $c_i$:

$$\hat{V}_i = \hat{p}_i \hat{a}_i r_i - c_i.$$

That extension is not evaluated in the present paper because controlled treatment-response data are not yet available.

Predictions are exposed through an interactive portfolio dashboard with segment filters and client drilldowns. A conversational agent further allows authorized users to ask natural-language questions about model outputs, portfolio trends, and individual predictions. The agent and dashboard retrieve prediction results from an internal Fast API service. The system is advisory: it does not deny services, block transactions, or initiate automated client action.

### 3.9 Evaluation protocol

Stage 1 is evaluated primarily with PR-AUC, threshold precision, recall, F1, and capacity-constrained lift. At review fraction $k$, lift is

$$\text{Lift@}k = \frac{\text{Precision@}k}{\pi},$$

where $\pi$ is the observed positive rate. Capture is the fraction of all positives contained in the top-ranked $k$ of observations.

Stage 2 is evaluated using RMSE and $R^2$ on the log-transformed target, MAE on the original dollar scale, amount-bucket accuracy, within-one-bucket accuracy, and threshold-specific precision and recall for large outflows. Stage 3 and Stage 4 are evaluated using accuracy, weighted F1, macro-F1, per-class precision, per-class recall, and confusion matrices. Macro-F1 receives particular emphasis because it gives equal weight to minority classes.

No separate probability-calibration study was completed, so Stage 1 scores should be interpreted primarily as ranking and relative-risk measures.

## 4 Results

### 4.1 Stage 1 discrimination and threshold performance

The Stage 1 classifier produced PR-AUC values of 0.828 on training, 0.823 on validation, and 0.823 on test.

At the validation-selected threshold $\tau^* = 0.3321$, the test confusion matrix is

$$\begin{bmatrix} \text{TN} & \text{FP} \\ \text{FN} & \text{TP} \end{bmatrix} = \begin{bmatrix} 140{,}791 & 935 \\ 1{,}969 & 5{,}924 \end{bmatrix}.$$

**Table 4. Stage 1 test performance at the validation-selected threshold**

| Metric | Value |
|---|---|
| PR-AUC | 0.823 |

| Metric | Value |
|---|---|
| Precision | 86.4% |
| Recall | 75.1% |
| F1 | 0.803 |
| Accuracy | 98.1% |
| Specificity | 99.34% |
| Balanced accuracy | 87.2% |
| False-positive rate | 0.66% |

Accuracy is high because negatives constitute nearly 95% of the test set; the more relevant finding is that the model retrieves three quarters of observed fragmentation cases while 86% of its positive flags are correct.

## 4.2 Capacity-constrained ranking and lift

The ranking results are particularly strong from an operational perspective. Among the top 1% of test observations, 95.3% are true fragmentation cases. Expanding review to the top 5% captures 78.7% of all observed positives.

**Table 5. Stage 1 lift analysis**

| Review capacity | Clients reviewed | Precision | Capture of all positives | Lift |
|---|---|---|---|---|
| Top 1% | 1,497 | 95.3% | 18.3% | 18.3x |
| Top 2% | 2,995 | 94.9% | 36.4% | 18.2x |
| Top 5% | 7,488 | 82.0% | 78.7% | 15.8x |

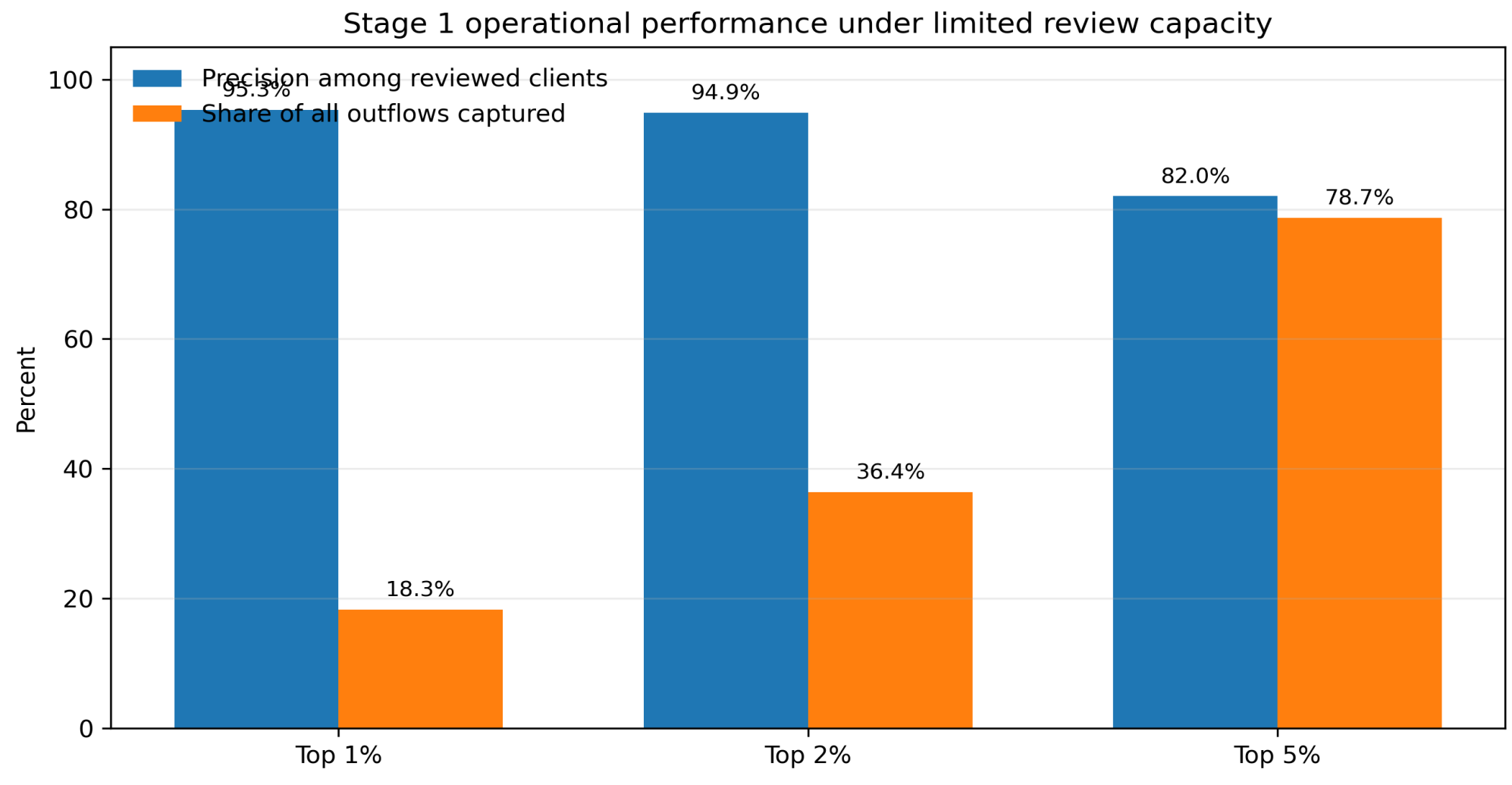

The top 5% of the test set contains nearly four fifths of all positive cases.

These results support the choice of a ranked decision-support workflow. A relationship-management team that can review only a small fraction of the portfolio can focus on a list with substantially higher event prevalence than the population base rate.

## 4.3 Stage 2 amount estimation

The amount model is evaluated on 7,893 positive test observations. It achieves a log-scale RMSE of 1.3613 and $R^2$ of 0.4998. On the original dollar scale, MAE is $11,098.

**Table 6. Amount-model test results**

| Measure | Result |
|---|---|
| Positive test observations | 7,893 |
| RMSE on $\log(1 + a)$ | 1.3613 |
| $R^2$ on $\log(1 + a)$ | 0.4998 |
| MAE on original dollar scale | $11,098 |
| Exact bucket accuracy | 62.6% |
| Within one adjacent bucket | 94.9% |

The bucket report shows the strongest F1 for the smallest two ranges and weaker recall for the largest range.

**Table 7. Amount-bucket test performance**

| Actual amount bucket | Precision | Recall | F1 | Support |
|---|---|---|---|---|
| $0-$500 | 0.86 | 0.59 | 0.70 | 2,410 |
| $500-$2,500 | 0.60 | 0.74 | 0.66 | 2,807 |
| $2,500-$12,500 | 0.49 | 0.62 | 0.54 | 1,735 |
| $12,500+ | 0.65 | 0.39 | 0.49 | 941 |
| Macro average | 0.65 | 0.58 | 0.60 | 7,893 |
| Weighted average | 0.66 | 0.63 | 0.63 | 7,893 |

Threshold analysis further reveals the high-value limitation.

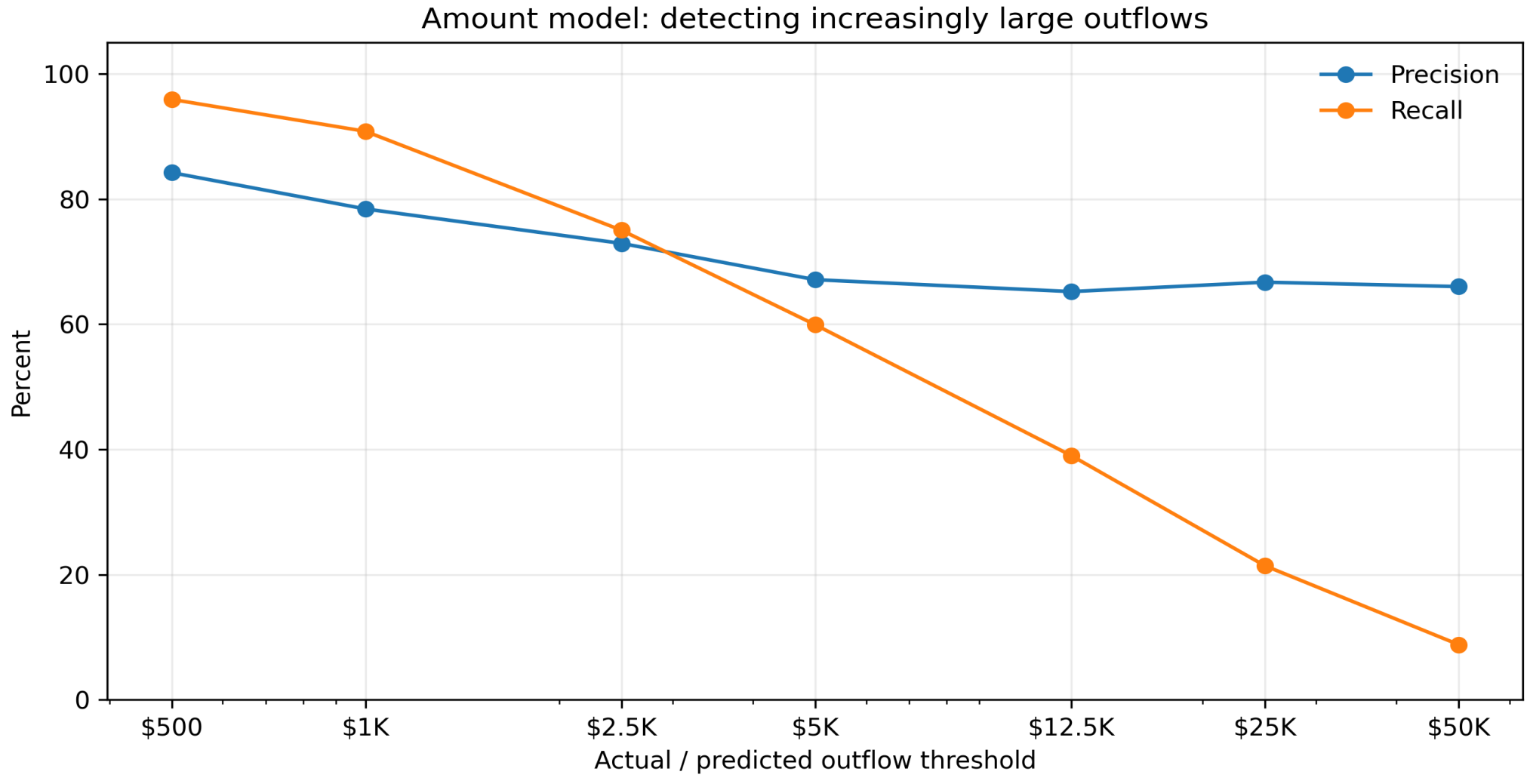

Precision remains comparatively stable as the dollar threshold rises but recall falls sharply for very large outflows. This result is important for interpretation. The model provides useful continuous ranking and broad amount segmentation, but it should not be presented as a precise estimator of the most extreme tail. The project report's high-value predicted-to-actual ratio of 0.64 is consistent with this underprediction.

## 4.4 Stage 3 originating-product prediction

The six-class source-product problem is imbalanced. Chequing accounts for roughly 91% of positive training examples, while the GIC class contains only a handful of observations. On the test set, overall accuracy and weighted F1 are both 0.92, but macro-F1 is 0.49. High aggregate accuracy is driven by the majority chequing class since we don't track multi-hop movement. Minority products remain materially harder to recover. The source model is therefore useful for the dominant chequing pathway and moderately informative for savings and mutual funds, but it is not yet reliable for rare product classes, unless more data is added.

### 4.5 Stage 4 destination-institution prediction

The destination model classifies 27 normalized external institutions and achieves 85% test accuracy, macro-F1 of 0.81, and weighted F1 of 0.85. The closeness of macro and weighted F1 indicates substantially better cross-class balance than the originating-product model.

**Table 8. Destination-model aggregate results**

| Metric | Train | Validation | Test |
|---|---|---|---|
| Accuracy | 0.90 | 0.86 | 0.85 |
| Macro-F1 | 0.91 | 0.81 | 0.81 |
| Weighted F1 | 0.91 | 0.86 | 0.85 |

Selected class-level results show strong performance for the most common digital platforms and several major banks, with lower scores for smaller-support destinations. The destination model is the strongest of the conditional classification stages. Its output can support competitor-level portfolio analysis and more context-aware conversations, while rare-class uncertainty should remain visible to users.

### 4.6 Overall empirical picture

Across the cascade, the evidence is strongest for *who* is likely to fragment and *where* the funds are likely to go. The Stage 1 ranking sharply concentrates positives within the top of the portfolio, and destination prediction remains strong across a broad class set. Amount estimation captures meaningful structure but loses recall in the high-value tail. Source-product prediction is accurate for chequing but weak for rare products. These differences matter because a multi-stage system should not present every output with the same level of certainty.

## 5 Discussion

### 5.1 From churn prediction to early-warning decision intelligence

The Stage 1 results support the central premise of the project: external flow behavior can provide a high-signal pre-churn target. A test PR-AUC of 0.823 and the top-ranked results show that the signal is operationally concentrated. The top 1% list is not merely somewhat better than random; it contains true positives at a reported 95.3% precision. Reviewing the top 5% captures nearly four fifths of all observed fragmentation cases.

This concentration is what converts a predictive model into a usable retention workflow. A bank does not need every client assigned a binary intervention. It needs a credible order in which to investigate relationships when staff capacity is constrained. The expected-outflow score adds a second dimension by distinguishing high-probability small movements from somewhat lower-probability large movements. This follows the broader insight from profit-driven churn research that the highest predictive score is not necessarily the highest-value action (Verbeke et al., 2012).

The design also moves beyond a single opaque risk score. Destination prediction can reveal whether risk is concentrated toward a digital investment platform, a direct bank, or a traditional competitor. Source-product prediction can indicate where the relationship is exposed. The dashboard and conversational layer make these distinctions accessible to non-modeling users.

### 5.2 Behavioral interpretation

The dominant importance of destination diversity over three and six months provides a compelling behavioral interpretation. Clients who send funds to multiple external institutions appear more likely to continue doing so than clients whose activity is absent or concentrated. Recent acceleration and source-product history provide additional signal. These features describe a progression: exploration of several providers, repetition of external activity, and increasing momentum.

The finding should not be translated into causal language. A high count of destination institutions does not itself cause fragmentation, and an intervention aimed solely at reducing that count would be nonsensical. Rather, it is a compact marker of relationship diversification. The most valuable human use is diagnostic: a relationship manager can investigate whether the pattern reflects pricing, digital experience, advice needs, investment preferences, or a legitimate multi-bank strategy.

## 5.3 Strengths of the modeling design

Several design choices increase confidence in the reported signal. The label is explicitly forward-looking, features end at the observation month, and a three-month buffer separates training from the evaluation period. PR-AUC and top-k measures reflect the rare-event and capacity-constrained setting more directly than accuracy. Value-aware weights and the expected-outflow score align modeling with economic relevance without requiring unverifiable assumptions about treatment response. The use of the same modeling family across stages simplifies deployment, reproducibility, and model maintenance.

The cascade also avoids forcing a single model to solve incompatible targets. Binary classification, regression, and multiclass prediction have different losses and error structures. Separate models allow the system to optimize each question and expose different measures of uncertainty. The amount results, for example, can be described as broad magnitude guidance rather than falsely precise dollars, while the destination output can be interpreted through class probabilities.

## 5.4 Limitations

All results are based on a 10% sample. The sample is large enough to train the current models, but rare products and small destination classes remain data constrained. Full-scale retraining may materially improve minority-class performance. It may also introduce new computational and feature-pruning requirements.

Second, the model detects direct external flows. Multi-hop movements, such as a transfer through an intermediary before reaching the final destination, are not observable as a single destination path. Destination labels are only as accurate as the transaction normalization and umbrella mapping.

Third, the amount model underpredicts the extreme tail. The sharp decline in recall above $12,500 indicates that the current log-MSE design favors the center of the distribution. Quantile regression, two-part frequency-severity models, or a dedicated high-value expert model should be evaluated.

Finally, feature importance is associational. Model-derived rankings can reveal which variables help prediction, but they do not identify the intervention that would change an outcome. Retention strategy should therefore combine model evidence with client context, product expertise, and, eventually, causal or uplift evaluation.

## 5.5 Governance and responsible use

The system is designed as advisory decision intelligence rather than an automated decision-maker. Predictions are shown to portfolio, growth, and retention teams that apply professional judgment before any outreach. The model does not deny services, block transactions, change pricing, or execute account actions. This boundary reduces the direct harm associated with false positives.

False negatives remain economically important because an actual outflow may occur without proactive engagement. The top 5% capture of 78.7% indicates that a capacity-limited workflow can identify most, but not all, positive cases. Users should understand that absence from the priority list is not evidence that a client will not fragment.

Access to raw data and prediction outputs is restricted within the enterprise environment. Personally identifiable information is encrypted and role controlled. Model artefacts, feature logic, class mappings, and random seeds are version controlled. Productionization should add drift monitoring, monthly prevalence and lift tracking, segment-level performance review, prediction expiry, and a rollback mechanism. These practices align with broader recommendations for reducing hidden machine-learning technical debt and documenting intended use (Mitchell et al., 2019; Sculley et al., 2015).

## 5.6 Future research and development

The next empirical priority is model development to target the extreme amount tail. Candidate approaches include quantile objectives, Tweedie regression, mixture models, and a two-stage high-value classifier plus regressor.

The most important long-term extension is treatment-effect measurement. The present system predicts who is likely to move money, not who is likely to be retained *because of* an intervention. Randomized or carefully designed quasi-experimental outreach data would enable uplift modeling and replace the current score with expected retained value. That step would close the loop between risk prediction and intervention effectiveness.

# 6 Conclusion

This paper introduces financial fragmentation as an actionable pre-churn prediction problem and demonstrates a four-stage temporal XGBoost framework for identifying, quantifying, and contextualizing external fund migration in retail banking. Using 595,220 client-month observations and 346 engineered features, the system achieves strong rare-event discrimination and highly concentrated top-of-list performance. The Stage 1 classifier reaches a test PR-AUC of 0.823, 86.4% precision, and 75.1% recall at the validation-selected threshold; the top 5% of ranked clients captures 78.7% of observed positive cases. The destination model performs strongly across 27 classes, while the amount and source-product results reveal clear boundaries: very large, rare outflows and rare originating products remain difficult.

The central contribution is not a claim that a model can determine why a client moves assets or automatically decide how the bank should respond. It is a practical decision-intelligence layer that moves the organization earlier in the relationship lifecycle, from reacting to completed churn toward investigating emerging fragmentation. The most responsible path forward is to preserve that human-review boundary while strengthening chronological evaluation, uncertainty measurement, minority-class performance, and causal treatment-response evidence.